# A Framework for learning multi-agent dynamic formation strategy in real-time applications

Mehrab Norouzitallab[a], Valiallah Monajjemi[b], Saeed Shiry Ghidary[c], Mohammad Bagher Menhaj[d]

[a] norouzitallab@aut.ac.ir    [b] monajjemi@aut.ac.ir    [c] shiry@aut.ac.ir    [d] tmenhaj@ieee.org

[a, c] *Department of Computer Engineering and Information Technology,*

*Amirkabir University of Technology, Tehran, Iran*

[b, d] *Department of Electrical Engineering,*

*Amirkabir University of Technology, Tehran, Iran*

**Abstract.** Formation strategy is one of the most important parts of many multi-agent systems with many applications in real world problems. In this paper, a framework for learning this task in a limited domain (restricted environment) is proposed. In this framework, agents learn either directly by observing an expert's behavior or indirectly by observing other agents' or objects' behavior. First, a group of algorithms for learning formation strategy based on limited features will be presented. Due to distributed and complex nature of many multi-agent systems, it is impossible to include all features directly in the learning process; thus, a modular scheme is proposed in order to reduce the number of features. In this method, some important features have indirect influence in learning instead of directly involving them as input features. This framework has the ability to dynamically assign a group of positions to a group of agents to improve system's performance. In addition, it can change the formation strategy when the context changes. Finally, this framework is able to automatically produce many complex and flexible formation strategy algorithms without directly involving an expert to present and implement such complex algorithms.

## 1. Introduction

Formation Strategy is one of the state of the art research topics in multi-agent systems attracted many attentions [1-8]. In formation strategy, an agent tries to find a proper position for itself based on the environment's state in order to increase its performance. Concerning a set of features that defines the environment's state, an agent tries to decrease the error between its current and the desired position to increase its cooperation with other agents. In other words, formation strategy for each agent can be viewed as a function whose inputs are some features elicited from the environment and its output is a target position for the agent. During execution time, agents try to get close to their target positions.

Formation Strategy is used in many applications, such as behavior analysis of fishes [9] and birds [10] and capturing/enclosing an invader [15]. It is also being used in many systems such as autonomous highway systems [8] satellites and spacecraft [11-13], UAV and AUV [14].

Depending on the problem, different models for Formation Strategy have been developed. Some of the most famous models are "Leader Follower" [3, 7, and 16], "Behavior Based Model" [17, 18], "Virtual Structure" [19] and "Model Based Formation Strategy" [20].

In the leader follower model, a leader is being followed by a group of agents. The main problem of this approach occurs when the leader suffers from a fault, thus generating inappropriate and weak features. Neglecting this drawback, this approach is widely used because of its simple and extensible nature.

In the behavior based model, a set of default behaviors are defined for each agent. The agent assigns some degree of importance to each behavior during execution, and then selects an action consistent with the most important behavior. Analytical formulation and theoretical analysis for this approach is difficult, so convergence to optimal configuration is not guaranteed.

In model based formation strategy, there are well-defined analytical models for the agent, tasks and the environment. In addition to time and effort needed to generate these models, the performance of the model is limited to known environments. In comparison to behavior based model, in this method there is no need to extract a model during execution. Furthermore, a goal is defined for each behavior and the entire task can be covered by the all behaviors.

Formation strategy based on virtual structure uses a fixed model. In this method, tasks are not individually assigned to agents; instead the whole formation strategy is presented to the all agents. Using this approach, the agent's behavior is predictable; as a result, agent's formation will be appropriate. One of the disadvantages of this method is the high bandwidth required for communication among agents.

In multi-agent systems, formation strategy function should possess some qualities such as smoothness of a sequence of target positions in a period of execution time. Moreover, being robust to the errors in input features is vitally important. Perception noise and partial observability of the environment are two prevalent causes of the errors. Furthermore, the most important quality of a formation strategy in a multi-agent is its responsibility of organizing agents in the environment in such a way that their consistency and coalition become guaranteed. Beside, this organization should prime a basis for agents in order to improve their behaviors' performance.

In this paper a framework is proposed to address diverse variety aspects of generating an appropriate formation strategy using supervised learning algorithms in many multi-agent systems with restricted environments . The main parts of this framework are Context Based Reasoning, Modular Scheme, Formation Strategy Function and Dynamic Position Assignment. Each part is responsible for solving a distinct problem concerned in a proper formation strategy.

Context based reasoning [21] allows agents adapt their formation strategy to the variations in environment's contexts, like humans who can make decisions based on a limited amount of data in different situations.

Formation strategy is an operation which the agents must be organized to perform their task. Indeed, there is a well-defined goal which is broken down into simpler sub-goals for different contexts. A hierarchical structure is proposed to reach these sub-goals. This structure is a powerful tool which eases the process of dividing a complex task into manageable sub tasks. First, the environmental status determines the context of the environment in each moment of the agents' execution time. For each context, an organization is conducted for agents by modular scheme. For example, agents with similar duties are grouped together. Agents in each group can learn their formation strategy based on local features of the environment and their leaders' position. By organizing the leaders based on global features of the environment, the coordination can be improved. In this way, the local features of the environment are directly presented to the formation strategy function of each agent and the global features are propagated through leaders' positions. It is clear that a high consistency among agents as a whole is feasible in this framework.

The framework emphasizes the application of machine learning to produce a complex formation strategy functions with ease. To this end, there are many difficulties concerned in using machine learning algorithms. The high coalition among agents causes the machine learning approaches to confront difficulties. Therefore, modular scheme is proposed to overcome such problems beside its application to agents' organization.

Considering the characteristics of the formation strategy problem, developing a general approach which could be used in different environments is of great interest. Machine learning is one of the methods that can be used to overcome this problem. Using machine learning, an agent is able to learn formation strategy, without the need for using any particular algorithms to reflect different requisites of each environment separately. This can be conducted either by using data generated by an expert, or data that is obtained by the agent itself via interaction with the environment. In fact, using machine learning methods, the expert can teach the agent complex and efficient behaviors via demonstration, rather than using complex algorithms.

In machine learning, patterns and features are extracted from experimental data. Learning from observation [22] is one of the methods to supply required data needed in machine learning problems. In this method, data are gathered by observing an expert's actions; next a learning algorithm is applied to extract expert's behaviors from the data. The learning phase is conducted according to other parts of the framework and especially modular scheme. Moreover, some regulations are proposed in order to specify the form that the data must be gathered and presented to the learning algorithm. These regulations influence the result formation strategy.

Many multi-agent systems in real world problems are complex and under burden of many details. Considering all their features or even effective ones (which are usually many) is impossible in practice. First, large amount of data are needed for learning which consequently increases the learning time. Second, problems like agents' position instability, features' redundancy, poor coordination, perception noise, and lack of enough information about a large amount of features will occur due to partial observabilty.

The modular scheme is crucially important in the framework, because it is responsible for organizing agents as well as eliminating problems concerned in the learning process. By using the modular scheme, the most important features of the environment are considered either directly or indirectly in formation strategy. The modular scheme is itself a hierarchy. Each group of agents is broken down into smaller groups and a leader (or some local features) is selected for each group. While other agents try to learn to follow their group's leader, leaders themselves are coordinated by following a common leader or other global features of the environment. The modular scheme is efficiently flexible as agents can adapt their formation strategy according to both a leader and a set of environmental features simultaneously. These features are considered in group levels. This structure increases both the intra-group and inter-group coordination as well as the learning accuracy.

In order to use the framework, each agent must first find its relevant section of the hierarchical structure. By using the local or global features of that section, it should then find its goal position. Finding the relevant section of the hierarchy, which is usually the role of the agent, can be done either statically or by dynamic position assignment method. In the latter case, a group of agents can choose their target positions from a group of proper positions in a way that brings maximum profit for the whole system.

In sum, the formation strategy framework has different parts with different duties. First, the formation strategy task is divided into some context. In each context, modular scheme organizes agents and prime a structure for learning the formation strategy function in each context. Each agent learns its formation strategy based on experimental data which is input-output features specified by the modular scheme. During execution time, agents determine the context of the environment and its related modular scheme. Then, agents present the value of features specified by the modular scheme to their learned formation strategy function as inputs to obtain their target positions. If dynamic position assignment method is active, agents can estimate the target positions of some other agents (if the other agents have the same ability) and choose their proper target positions to increase the system's performance. Dynamic position assignment method has some characteristics to avoid conflicts among agents even in a partial observable environment.

The remainder of this paper is organized as follows. Section two describes main parts of the framework. In section three, Context Based Reasoning is reviewed while section four is devoted to modular scheme's explanations. In Section five,

Dynamic Position Assignment method is proposed. Experiments are presented in Section six and finally section seven concludes the paper.

## 2. A framework for leaning formation strategy

Methods like formation strategy function, modular scheme, context based reasoning, and dynamic position assignment, each represents a distinct set of features for the problem of formation strategy. Integrating all these features in a framework in order to be used in multi agent systems in restricted environment can be significantly considerable. This integration is obtained in the framework by a hierarchical structure.

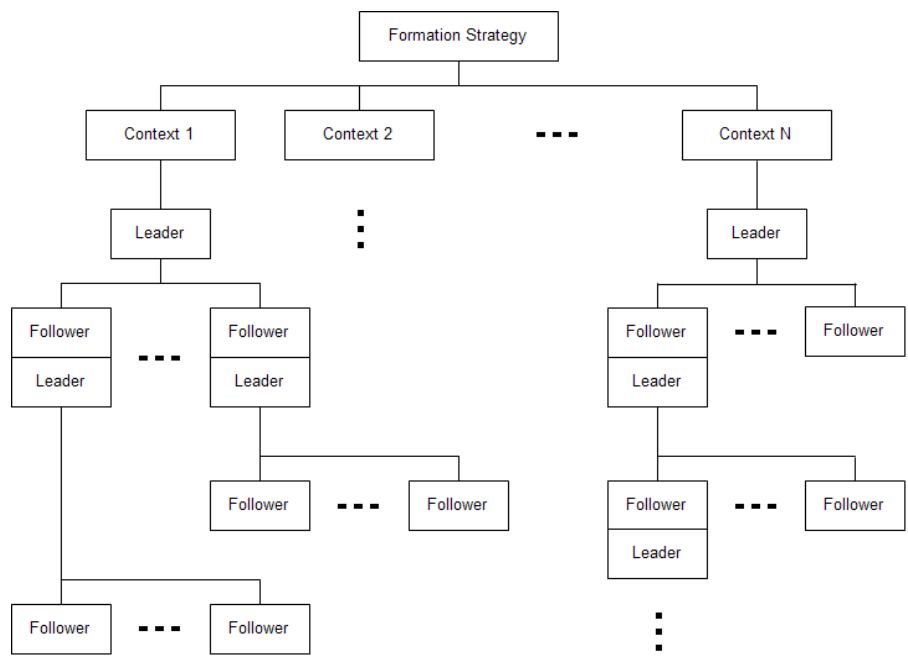

Figure 1. Hierarchical Structure

In figure 1, a diagram of the hierarchical scheme for the formation strategy task is depicted. In this method, formation strategy task is divided into some contexts; each of them contains a hierarchical set related to its modular scheme. In modular scheme's hierarchical structure, each leaf represents an agent, which follows a leader or a set of features (if exists). With the help of its formation strategy function for specific context and based on a set of features, each agent finds its target position. In case that some agents know the other's proper positions or they want to choose their positions from a group of candidate positions, while avoiding interference and increasing system performance, the dynamic position assignment method is used. The dynamic position assignment method substitutes optimally the roles of agents according to their target positions during the execution time.

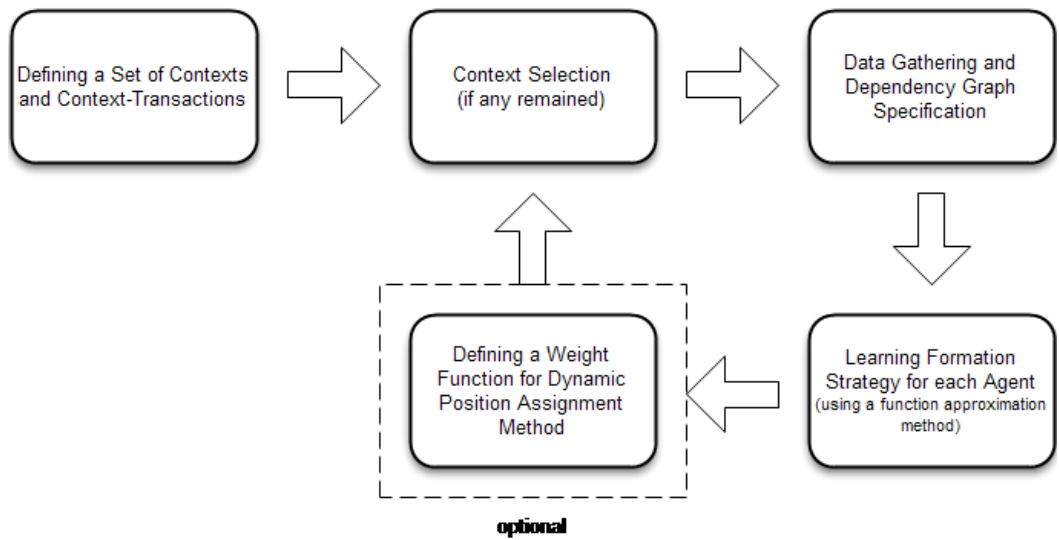

Figure 2. The framework's block diagram for formation strategy

Figure 2 shows the framework for learning formation strategy. According to this diagram, the expert first defines some contexts as well as transitions between them. Then he defines a hierarchical structure for each context by modular scheme. Generating a formation strategy algorithm according to this structure needs data gathering as well as specifying dependency graph which is a representation of modular scheme. In the next step, the formation strategy function for each context is approximated using a supervised learning method like neural networks. This approximation is based on gathered data and the dependency graph. In order to use the dynamic position assignment method, weighting each agent-location pair to specify the assignment performance is vital.

It is apparent that the expert does not need to implement the formation strategy directly. This framework is general enough to be used in many multi-agent systems in restricted environments, therefore by developing a graphical user interface or a toolbox; experts without explicit knowledge about implementation methods can apply this approach.

## 3. Context Based Reasoning

Context based reasoning allows the agents to exhibit high level behaviors. This reasoning is based on the idea [38] that:

1. Context is a set of actions and procedures that properly addresses the current environment's conditions.

2. As a task evolves, a transition to another set of actions and procedures is required to address a new context.

3. Everything happens in a contexts is influenced by and limited to the same context.

Agents can perform predefined tasks using context based reasoning. For each task, some goals are defined for each agent. There are limitations to reach each goal [38]. Each context represents a state, based on environment's status and agent's motives. This state causes a specific behavior for each agent in that context [38].

The main three concepts of context based reasoning are task, context and transaction function.

A task is an abstraction defined inside a model which is set for an agent prior to runtime. A task consists of a target, a set of constrains and a context that dictates high level behaviors to the agent. The target defines the conditions in which the task terminates. The target can be well defined by environment's (physical or environmental) conditions. Limitations are physical, logical and environmental.

In [38], physical and environmental conditions that represent a specific behavior are considered to comprise a context. Context based reasoning's model requires that agents to be active during each time-step. A context in the model is said to be active when the condition implying its validity is satisfied and the agent uses its own knowledge to make decisions for its task.

The task describes an agent's high level behavior by defining groups of context-context transitions pairs. Context transition denotes specific changes between contexts during the task's execution.

A set of contexts $C = \{c_1, c_2, ..., c_n\}$ are considered for a task. The role of an expert is to design a model by defining contexts, consistent with his intuition about the task. Each context needs to be paired with a behavior in order to perform a task. This technique is more appropriate for tasks that agent's behaviors and activities are well defined. In addition, this technique is useful when each task execution is dependent on a series of sub-tasks.

## 4. Learning Formation Strategy Using the Proposed Modular Scheme

### 4.1 Definition of Learning Formation Strategy

Assume that $A = \{a_1, a_2, ..., a_n\}$ is a set of attributes of the environment. $V_t = \{v_1^t, v_2^t, ..., v_n^t\}$ is a set consisting of the values that represent the degree of similarity of the environment at time $t$ to set $A$. If the environment is partially observable at $t$, some of $V_t$ values are not measurable and thus are considered unknown.

A history of measured values of attributes for k time-steps is defined as $hist_v = \{V_{t-k}, V_{t-k+1}, V_{t-k+2}, \ldots, V_t\}$. $m$ number of features that have influence on formation strategy are extracted using function $\varphi: hist_v \rightarrow P$ in form of set $P = \{p_1, p_2, p_3, \ldots, p_m\}$. Formation strategy for an agent is defined as function $\delta: P \rightarrow T$ where $T$ is agents' target positions in their formation strategy.

In learning formation strategy, the function is approximated using data gathered by an expert. In this paper, supervised learning methods are used for function approximation. According to the definition, the output of the function ($T$) is the target position for the agent. It is assumed that there exists a motion planning method to reach the target position. The motion planning can also be learned using other methods which are not covered in this paper.

### 4.2. Modular Scheme

One of the important issues in learning the formation strategy in multi agent environments is the features and their relations to each other. The changes made by each agent in the environment when executing its formation strategy, have direct influence on features spotted by other agents. Therefore, the agent's formation strategy has to be defined in a way that can maintain the system's coordination. There are several concerns about learning an effective formation strategy.

- **Coordination reduction among agents:** If in learning formation strategy, there were no features defined to increase agents' coordination, each agent would use environment's local and global features, without considering the performance of other cooperating agents, in order to improve the performance of its own formation strategy. This solo improvement may cause a dramatic drop in performance of cooperative agents, thus the whole system. Considering the coordination in formation strategy, will cause the agents to support each other when executing tasks.
- **Instability:** Consider two agents, agent 1 and 2. If agent 1 observes the position of agent 2 and agent 2 observes the position of agent 1, either directly or indirectly, with every move of agent 1, there will be a change in target position of agent 2 and vice versa. The dependency between two agents will cause instability in agents' positions until one agent stops moving.
- **Lack of proper information about features during run-time:** Observation in many multi agent systems are partial, in addition, communication between agents are costly. As a result, when the number of features which have direct influence on formation strategy increases, some of those features may not be available or with much error during run-time.

    **Feature redundancy:** If a set of features can be extracted from another set, considering both in a single set directly in learning will cause extra overhead during run-time. In addition, if an agent receives each similar set with different delay from different source, an additional noise is not

avoidable. For example, in a partially observable environment, this extra overhead will force the agent to extract information from both sets, nevertheless only one feature set is adequate. Considering three agents, agent 1, 2 and 3, If agent 2 follows agent 1 and agent 3 wants to consider the features used by agent 1 and 2, the position of agent 2 will provide enough information about those features because the feature set used by agent 1 will be propagated to agent 3 through agent 2 Taking into account the positions of both agent 1 and 2 or the other features will cause feature redundancy.

A modular scheme is proposed which can suitably overcome the mentioned issues. This modular scheme will be effective and useful in many multi agent systems. In this scheme, agents are grouped together based on some similarity factors. A leader is determined for each group, which other group members will follow its position. The leader will learn its formation strategy based on local and global features. Each agent will find its proper position based on its leader position. Considering common global features in learning formation strategy for leaders, increases the coordination between groups. In addition, it is possible to use a leader who only uses system's global features in order to coordinate the whole system. This modular scheme causes the transfer of prominent features from leaders to their follower agents. Furthermore, each group's agents are coordinated by following their common leader, which consequently increases the cooperation among them. The presence of a leader, gives agents the ability to consider the leader's position as a local feature in their formation strategy to increase their coordination with other agents. Agents are not restricted to follow just a leader. Instead, agents of a group can consider a set of global features alongside a set of local features or even status of more than one agent in their formation strategy. Furthermore, the modular scheme is flexible enough to cover many models for formation strategy as well as leader-follower.

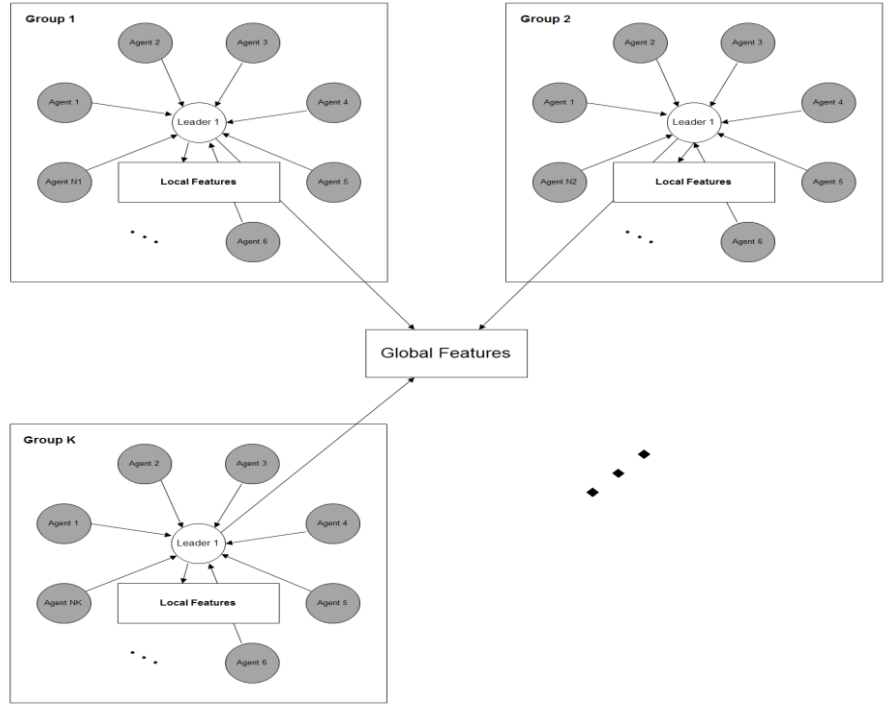

Figure 3. An overview of modular scheme alongside with the leader follower model

Figure 3 demonstrates the modular scheme's organization alongside with the leader-follower model. This combination transfers the attributes sensed by the leader to the agent with some delay proportional to the importance of the feature. During this transfer, other delays relevant to the perception process of the leader, like partial observation delay and the delay of communication between agents have less effect. In other words, if a global feature is passed through different leaders before it reaches an agent, it will cause more delay comparing to the features transferred from the main leader of an agent. Furthermore, when the number of leaders in the hierarchy increases or the error appears in a leader position - when it tries to reach its proper position - the transferred features to the agent will contain more errors.

Using machine learning methods with modular scheme based on a training data-set will help the agents in a group follow a leader using different methods when necessary.

In order to learn formation strategy with the help of modular scheme, it is needed that an expert gathers a training set. Then, this training set can be used to train a machine learning method like neural networks or support vector machines. In practice, this process can be time consuming in complex systems; therefore a systematic approach is proposed. In this approach, an expert gathers snapshots of the system, either manually or automatically. Each snapshot includes an instant set of features elicited from an environment status. Furthermore, the expert can manually determine the appropriate target position of each agent reflecting the environment status. When the goal is simulating or extracting an algorithm used by other agents or even creatures in real-world, the expert can be neglected in this process and the positions of the other agents (creatures) mirrors the target positions of the agents. The snapshots form a training set in a matrix form.Each row of the matrices represents a data related to a feature or a coordination of an agent's position. Each column is related to a snapshot. The expert must define the modular scheme as an acyclic directed graph, called the "dependency graph". Each node of this graph, stores the number of a specific row in the data matrix. Directed edges traveling outside of a node, the end node specifies the sources (features or other agents), from which the agent learns its formation strategy function. Thus, for each node representing an agent, a function approximation algorithm is used to learn the formation strategy function based on the knowledge represented by its dependencies and the training set.

The error of the estimator causes an extra error in formation strategy function learned by the agents following a leader. Assume that, each snapshot contains information about a group of features, the proper position of the leader and the proper position of a follower. In addition, the dependency graph contains a relation between the leader and the features as well as a relation between a follower and the leader. In order to reduce this extra error, it is better for the agent to use the estimated position of the leader - which contains the learning error- instead of leader's proper position existed in the data-set. The leader's proper position in the data-set is replaced by its estimated position during execution time. In order to solve this problem, first the dependency graph is topologically sorted [36] in reverse order (from features to agents). The agent's positions in the data matrices are then replaced by the outputs of the estimated formation strategy function, based on the input data, recursively.

### 4.3. Formation Strategy Function Approximation

To learn the formation strategy function, the expert must consider a set of features. Then, based on the selected features, he/she must gather data-set $D$ containing $(p,t)$ pairs. In this pair, $p$ is the feature vector of a specific state and $t$ represents the proper position of the agents for that state. The data could also be gathered via observing other agents or objects.

Different methods can be employed to approximate a given function based on the date set $D$. None of these methods are error free. We can define the approximation error of the formation strategy function as follows:

$$E = \frac{1}{N}\sum_{i=1}^{N} dist(t_i, \delta(p_i))$$

In this equation, N is the number of samples, $t_i$ is the $i^{th}$ coordination of t and dist is a function to measure distance between two points. $p_i$ is a set of features and $\delta$ is the approximated formation strategy function. If the points are defined in Euclidean space, the equation can be defined as follows

$$E = \frac{1}{N}\sum_{i=1}^{N}\sqrt{\sum_{j=1}^{d}\left(t_{ij} - \delta_j(p_i)\right)^2} \qquad eq.1$$

In this equation, d is the dimension of the environment, $t_{ij}$ is the $j^{th}$ coordination of the target position in $i^{th}$ datum sample and $\delta_j$ is the $j^{th}$ estimated coordination of the target position regarding $p_i$.

We can also define the "Sum Square Error" in this domain as follows:

$$\text{SSE} = \sum_{i=1}^{N}\sum_{j=1}^{d}(t_{ij} - \delta_j(p_i))^2 \qquad eq.2$$

It is easily apparent that equation (eq.1) is always lesser or equal than equation (eq.2).

As a result, the reduction in SSE will reduce the approximation error which appears in formation strategy function. A wide variety of learning methods can be used to reduce SSE.

## 5. Dynamic Position Assignment

In dynamic position assignment method, it is assumed that a set of agents Agents = {Agent$_1$, Agent$_2$, …, Agent$_n$} exists and each agent wants to choose a target position from the set P = { p$_1$, p$_2$, …, p$_k$} where $k \geq n$. In this method, the function $weight : i, j \in R \to R$ is defined to specify the appropriateness of assigning the position p$_j$ to Agent$_i$. This weight function is used by agent *Agent$_i$* for position selection. Then for each agent a distinct point will be assigned in a way that sum of weights for agent-position pairs become maximized.

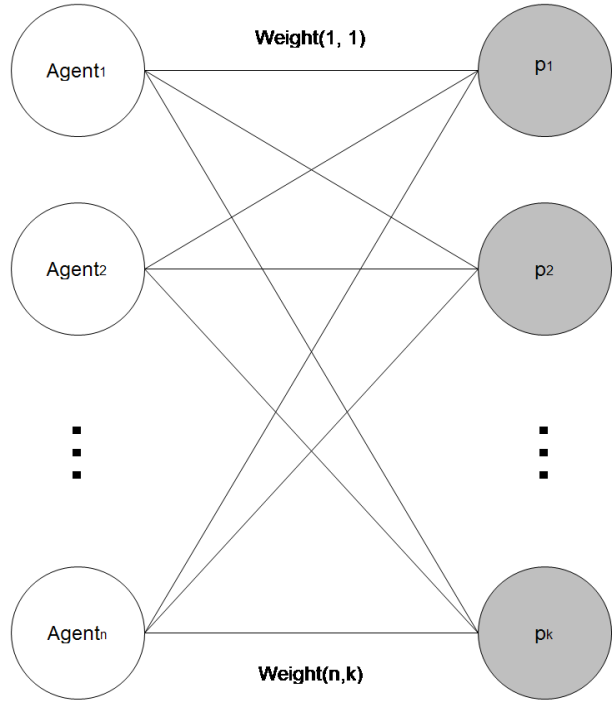

Figure 4. Dynamic Position Assignment's graph

According to figure 4, the problem can be modeled as a complete weighted bi-parted graph. The graph consists of two sets of independent nodes. $Node_{i,1}$ in the first set represents *Agents* while $Node_{i,2}$ in the second set represents $p_i$. The weight of each edge $Edge_{i,j} = \{Node_{i,1}, Node_{i,2}\}$ is determined by weight function.

The weight function is defined in an application dependent manner, for example in a multi robot system, the weight of each agent-position pair, can be determined based on some factors such as the distance between the agent and the position, agents' current battery (energy) status, agents' priority, agents' physical abilities, etc. These factors can be combined using arbitrary functions. Moreover, the weight function can be determined using machine learning or optimization techniques which are not covered in this paper.

In the next step, Hangarian algorithm [39] is used to match agent-positions in the graph in an optimized way. Matching in this problem means selecting a set of edges, each of them connects an agent to a unique position. Optimized matching (Maximum weighted matching) means selecting the set that maximizes the sum of assigned edges' weights. In other words, agent-position pairs are selected in a way that the system's performance becomes optimized based on weight function.

Dynamic position assignment is applicable in problems where agents of a group have some sort of approximation of other agents' formation positions in the group. Formation position of other agents can be estimated in various ways such as communication or using stored function approximation method of other agents. It is obvious that mentioned methods will cause some overloads and errors.

Using dynamic position assignment approach has significant improvements both on system's performance and agents' cooperation. Using this approach, if any agent's movement is changed toward a behavior other than formation strategy, its role in formation strategy can be changed dynamically with another agent in order to avoid inference between agents' tasks. In addition when an important agent such as a leader in leader-follower model, faces fault in a situation, its role can be substituted with another agent using dynamic position assignment approach. Moreover, this approach can reduce energy consumption in systems where agents have limited energy resources.

## 6. Experiments

In order to examine the proposed framework, we have used RoboCup soccer 2D simulation environment [24]. RoboCup soccer simulator 2D is a tool for research and education about multi-agent systems and artificial intelligence. This tool lets two teams of autonomous simulated agents to play soccer against each other and can be used to examine and compare different machine learning approaches in a complex multi-agent system.

The aim of a team in this simulated environment is to score more goals and receive fewer goals during a match. The formation strategy is one of the most important behaviors of each team's agents. Each agent must find its target position in a way that both its personal activities and its cooperation with other teammates improve.

The soccer simulator is a complex multi-agent system with many effective features. Objects in this environment are represented in two-dimensional Euclidean space. The field size is 105 x 68 meters.

The contexts in soccer simulation environment has been defined as "Attack", "Defense", "Mark", "Run Away" and "Dead Balls" (Throw-in, Corner Kick and Free Kick). Contexts' transitions are based on current game state and the active context. Then some data are generated for each context using a graphical tool by the expert. The formation strategy function for each context is approximated based on those data and a dependency graph.

Two separate data-sets have been generated by two different people with different field knowledge, one who was experienced expert in soccer simulation and one who was just familiar with that. Each set consists of some snapshots. Each snapshot contains the position of the ball and the proper position for eleven agents. These data has been generated using a graphical tool. The first set contains 955 snapshots and the second one contains 1000 snapshots.

In this paper, neural networks with Levenberg-Marquatre [31] back-prorogation algorithm is employed among applicable methods used to approximate the formation strategy function which reduces SSE function mentioned in section 4.3. In theory, a three layered neural network with sigmoid units in hidden layer and linear units in output layer is able to approximate every arbitrary function with reasonable error [30]. A neural network can extract expert knowledge from data to build a model with high generalization ability.

One of the main drawbacks of using a neural network is the risk of over-fitting on training set. In order to overcome this problem, we use early stopping method [40] in which, the training process will be stopped when the error is increasing on an independent validation data-set.

For training, data-set is divided into three sets, training, validation and test. Training set is used to train the neural network, Validation set is used for early stopping method and Test set is used to measure the generalize ability of the trained neural network. To this end, 70% of the data has been used for learning, 20% for testing and 10% for validation. The testing part of the each data-set was constant during and is used as an identical scale for error measurement in different experiments. A feed-forward neural network has been used to learn each mapping. The main focus was on the first data-set, the one which was generated by the expert.

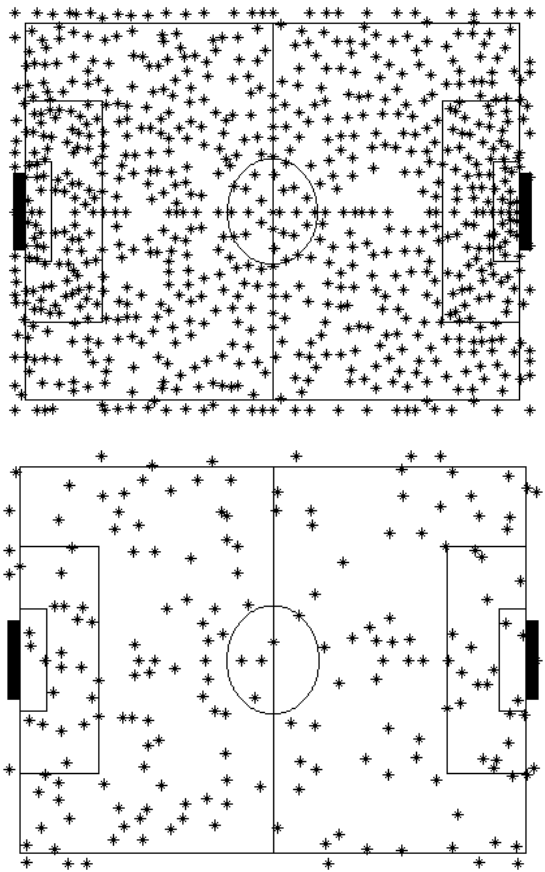

Figure 5. Positions of ball for training and test set

Figure 5 depicts the ball position in training (right) and test (left) portions of the expert's data-set.

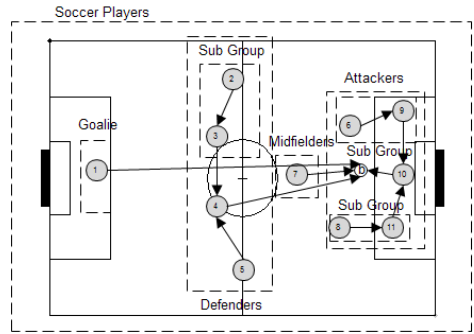

(a)

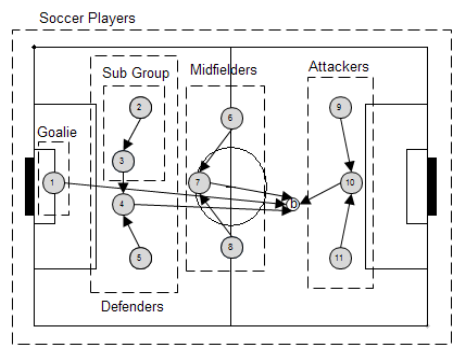

(b)

Figure 6. A view of modular scheme in a team of agents in Soccer Simulation environment in normal context (a) and attack context (b)

Figure 6 shows two different modular schemes in Normal and Attack contexts. This figure also shows the agents' groups and their dependency graph. It is possible to use more complex modular schemes in the simulation's environment based on local and global features in order to improve the system's performance. However, simpler schemes are used to better demonstrate the proposed approach's capabilities. Similar schemes have been generated for different context. In this section all experiments has been done using the Normal context. The experiments for other contexts shed similar results.

In the Normal context, agents have been divided into four groups: Goalie, Defenders, Midfielders and Attackers. Each group has a leader for coordination of agents. Leaders are coordinated based on the ball's position which is a global feature and also provides useful information about a game.

After creating the dependency graphs (Figure 6), agents' learning priorities is calculated using topological sorting. In this order (1-4-5-3-2-7-6-8-10-9-11), first agent number 1 learns its formation strategy based on ball's position from training data and then it updates training data. Next, agent number 4 as the defenders' leader learns its formation strategy in the same way. Then, agent number 3 learns its formation strategy function using the updated data-set (which includes its leader updated positions). This is done in order until the end of learning phase.

After the learning phase, the approach's accuracy has been tested using test portion of the data-set. For examining the test error, it is assumed that agents will reach their target position without considering transition delay and environmental features' value changes meanwhile the transition. First, the leaders find their proper position using the test data and topological order of the dependency graph based on the ball position and their approximated formation strategy function. Then, follower agents find their proper position based on their leader's position. The approximation error is then calculated based on the difference between the agents' position with their positions in test data. Using the expert's data-set, the formation strategy function based on modular scheme has been approximated with the error of 0.89 meters and standard deviation of 0.81 meters. The error has been increased by 5% when the data-set update after each learning step was omitted.

In order to examine the effect of using modular scheme and its stability, a scenario has been simulated. In this scenario, first a force is applied to a stationary ball, and then agents determine their formation strategy while the ball is moving. In each time step, each agent moves toward the proper position calculated using the formation strategy function with maximum speed of 0.5 meters /cycle (If the difference between its current position and the desired position is greater than 0.5 meters). The closest agent to the ball tries to approach the ball with maximum speed of 0.5 meters/cycle in each time step. This task added to the scenario to examine the coordination level of the framework when an agent executes a task independent to formation strategy.

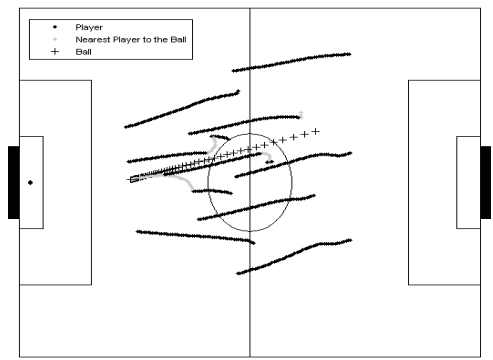

(a)

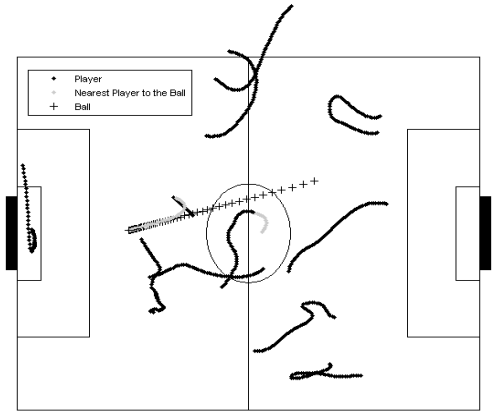

(b)

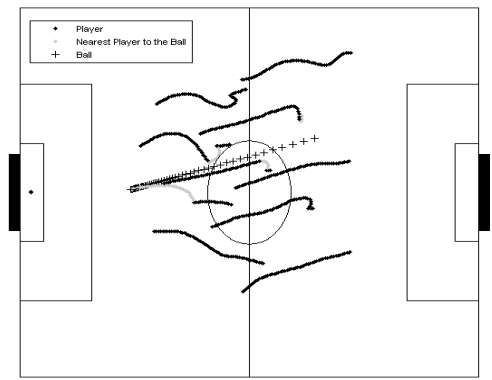

(c)

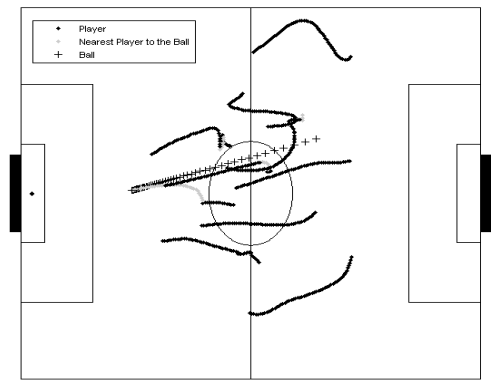

(d)

Figure 7. The output of learned formation strategy function for all agents according to

Only ball position (b) Ball position and teammates' positions

(c) Modular Scheme

(d) Presenting all the direct and indirect effective features of modular scheme directly

Figure 7 shows the trace of agents' and ball' position during the experiment. In (a) agents only used current ball position in learning their formation strategy. As it is shown in this figure, when each agent moves toward the ball, some vacant places created in different parts of the field, especially in defense, which adversely affects the team's overall aim of not receiving a goal. In (b) each agent uses a strategy function based on current position of the ball and all of other agents. This led to instability. (c) Shows the team's performance using modular scheme. As it is shown, the team's behavior is well coordinated via having some information about a leader's position(considering observation and communication delay will lead to a negligible more error). In (d) each agent uses a formation strategy function which involves all the features considered in modular scheme (even indirectly via a leader's) directly. As it is clear, agents' coordination and the quality of formation strategy under this strategy is lower comparing to the (c). For example, the bad performance of the left defender might be because of the absence of any relation between features (the ball and the other two defenders) as well as inadequate training data.

In fact, it can be concluded that, the modular scheme is helpful for learning formation strategy in soccer simulation environment. The resulting formation strategy is stable, coordinated and with good performance.

The dynamic position assignment has been used for the "Mark" context. Similar to [26] the learned formation strategy function for each player only determines positions where the opponent's movements can be under control. Other teammates mark other opponents using similar function. The resulting marking positions form the set needed in dynamic position assignment method. Some extra information like position of teammates (with some error) is also available to this method. A linear function as the method's weight function has been defined for each specific position (usually an opponent) using the value of features like the distance between the team's agents to the opponent position, distance between that specific position to the goal, distance between the position to the ball and the priority of the position (how much the position is dangerous). The weight of each factor in the aforementioned function is calculated using PSO optimization method in automated scenarios. This method resulted in a proper marking in which each agent marked an opponent without any inference with other agents. The most dangerous opponent was also marked in minimum time. All this performance has been obtained under the presence of noise in the environment, error in approximations and partial observation. Agents also were not being able to communicate with each other to share any information. Consequently, using dynamic position assignment in other contexts will cause the agents to swap their positions intelligently while doing tasks not defined in formation strategy, like intercepting a moving ball.

As a conclusion, using the proposed framework for the task of formation strategy in a team of agents in the soccer simulation environment resulted in a proper formation strategy and increased coordination among agents.

**6.1 Approximating Formation Strategy Function using Neural Networks**

In order to approximate the formation strategy functions with neural networks based on the features which are directly involving, the position of the ball which is represented by a couple $(X_B, Y_B)$ is used as input feature set. The position of the ball carries lots of information about the game state. Learning formation strategy function as a mapping from the position of the ball to target position for an agent can increase the coordination among agents, because the ball's position is a prominent global feature. Neglecting the fact that ball position is not a sufficient input feature for learning formation strategy as mentioned before, the analysis of the formation strategy function approximated by neural networks and a simple feature set like ball position sheds some interesting results. An approximation for formation strategy function according to another set of input features has similar results.

A separate neural network has been used for each agent to map the position of the ball to its target position using training data. In order to evaluate the approximation error, the approximation error of the formation strategy function defined in section 5 has been used. In addition the maximum error has been defined as the average maximum approximation error of all agents over the test data. For finding the proper neural network's structure, parameters and the size of data-set to train a neural network, some experiments are conducted considering special cases such as using early-stopping algorithm to avoid over-fitting. Therefore, the effect of increasing the number of neurons in the hidden layer and increasing the number of data on the average error and the maximum error are examined. In order to increase reliability, the results have been averaged over several experiments. The neural network structure and parameters have been kept constants for different agents to make this approach applicable to large environments with many agents.

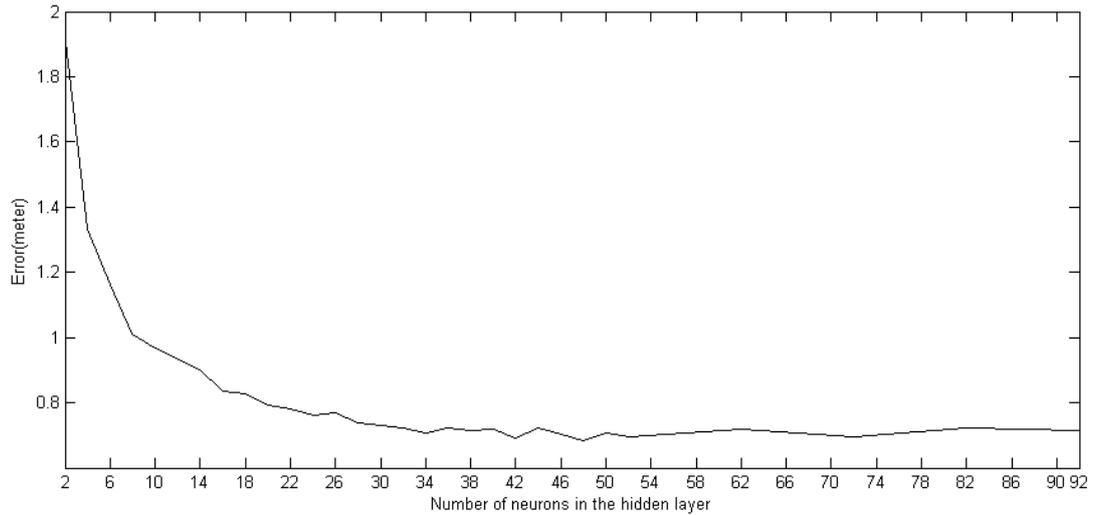

(a)

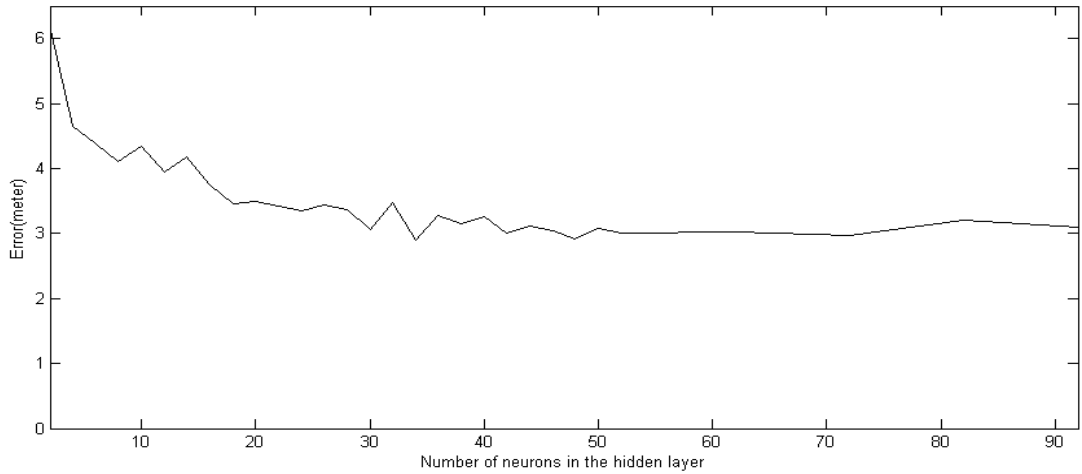

(b)

Figure 8. The effect of increasing the number of hidden layer's units on

Average error (b) Maximum error

In figure 8, the effect of increasing number of neurons in hidden layer on average and the maximum error on data-set 1 (provided by an experienced expert) is depicted. With the increase of number of neurons in hidden layer, the error first decreases, then remains steady after a certain point because of using early-stopping algorithm to avoid over-fitting. According to this figure, the proper number of neurons for the hidden layer is 36. When the data-set 2 used for learning the formation strategy function, the average and maximum error has been 300% and 670% more than the data-set 1 respectively. This shows the amount of error exists in this data-set which is not provided by an expert. data-set 2 has not been used in the following experiments.

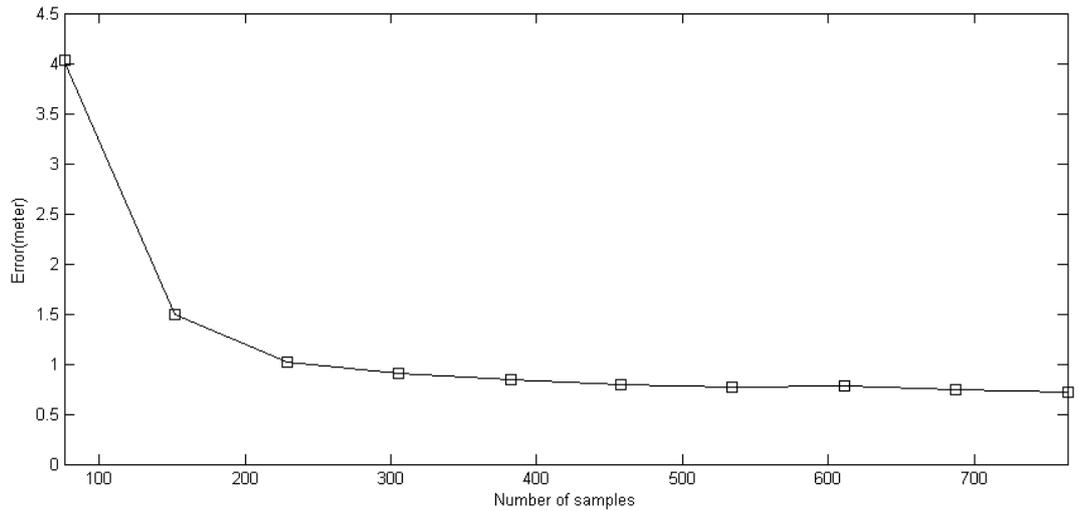

Figure 9. The effect of increasing the number of samples

Figure 9 shows the effect of increasing the number of samples in training data-set on learning formation strategy function. As it is depicted, the average error has been decreased with the increase in the number of samples. The slope of this decrease is low after a certain point. The maximum error follows a similar trend. According to figure 9, 764 samples of the training data are sufficient for approximating the formation strategy function. The average and maximum error using calculated optimized parameters has been 0.72 and 0.64 meters respectively.

In addition to error over test data, two other factors are important in approximating formation strategy function. The "smoothness" which increases while number of direction changes in each agent's movement's decreases and the robustness to noise which is the consequence of observation error in a non-deterministic environment.

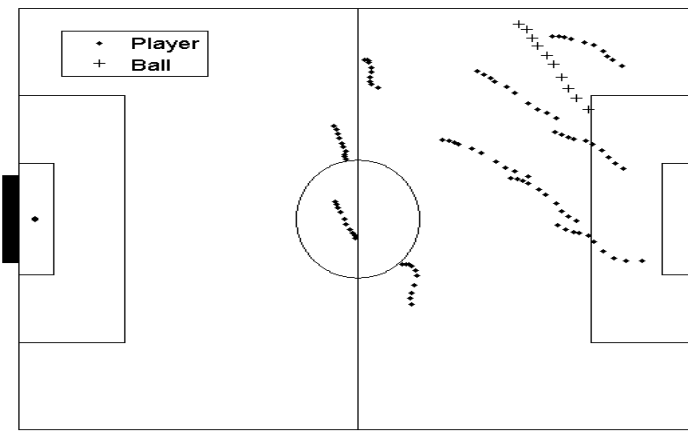

Figure 10. Examination of formation strategy's smoothness

Figure 10 shows the smoothness of the learned formation strategy function. In this experiment, the ball has been placed on a random initial position and started moving with a random velocity vector. While the ball was moving, agents followed their formation strategy function. In this experiment, a noise with variance of 0.15 meters has been added to the position of the ball in each cycle. To achieve better results, the experiments have been repeated for 100 times. The average change in agents' angle of movement during the experiment has been 12.62 degree with the standard deviation of 25.37 without adding noise to the ball's position. These values have been increased to 19.9 and 30.0 respectively after noise added to the ball's position. The average total distance that all agents moved in each cycle has been 0.64 meters with standard deviation of 0.53 meters without noise and 0.63 meters with standard deviation of 0.51 meters with noise. This shows the low effect of noise on the total distance. This experiment shows that the noise in position of the ball affects the angle of movement. It can be also stated that the high changes in angle of movement happened when the small distances has been traveled. According to these results, it can be seen that the resulting formation strategy function has a good smoothness in the soccer simulation environment. This is because of good training data, as well as continuity in the approximated formation strategy function.

Table 1. The effect of noise on average error and errors' variance

| Noise Variance | Average Error | Errors' Variance |
|---|---|---|
| 0.0 | 0.72 | 0.64 |
| 0.15 | 0.73 | 0.642 |
| 0.3 | 0.76 | 0.652 |
| 0.45 | 0.78 | 0.662 |
| 0.6 | 0.82 | 0.665 |

In order to analyze the robustness, a Gaussian noise with different variances has been added to position of the ball. Then the error of the resulting agents' positions has been compared to the target positions in the test data. This experiment can show the effect of noise on the output of the formation strategy function. Table 1 shows that the noise has a low effect on average error and the standard deviation. The experiments show that the resulting formation strategy is both smooth and robust to noise.

Using clustering with feed-forward neural networks for each cluster resulted in a performance similar to the networks without clustering. This method not only did not increase network's accuracy but also decreased the smoothness of the formation strategy function in boundaries. In addition, this method has more learning parameters which leads to slower learning speed, higher execution time and higher memory consumption.

In order to demonstrate another capability of the proposed method for formation strategy, learning has been done based on observing another set of agents. To achieve this, a set of training data has been prepared by observing the behavior of agents who followed SBSP algorithm [35] as their formation strategy. Their formation strategy function has been approximated with the error of 0.031 and standard deviation of 0.032 in a noise-free environment. It can be claimed that, the proposed method is able to learn different formation strategy methods.

## 7. Conclusion

In this paper, a framework for learning formation strategy was proposed. The framework possesses a diverse variety of features to address different issues associated in formation strategy behavior of agents in many multi-agent systems with restricted environment.

The four main parts of the framework are context based reasoning, modular scheme, dynamic position assignment method, and formation strategy function approximation. The integration of these parts conveniently generates a complex formation strategy algorithm for a multi-agent system.

Context based reasoning reflects the variations of environment status by changing its active context. Modular scheme provides an exhaustive organization of agents and primes the basis for learning formation strategy function. Formation strategy function approximation estimates the formation strategy function of agents according to a data-set and the modular scheme specifications. Dynamic position assignment method is able to dynamically specify the role of each agent during execution time to increase the performance of a system.

Finally, the framework was completely examined in a complex multi-agent system to demonstrate its capabilities. The results are very promising.

## 8. Acknowledgments

The authors thank Mohammad Mehdi Korjani for his valuable comments and sharing his knowledge.